\documentclass[runningheads]{llncs}

\usepackage{eccv}

\usepackage{placeins}
\usepackage{multirow} 
\usepackage{eccvabbrv}

\usepackage{graphicx}
\usepackage{booktabs}

\usepackage[accsupp]{axessibility}  % Improves PDF readability for those with disabilities.

\usepackage[breaklinks,colorlinks,citecolor=eccvblue]{hyperref}
\usepackage{orcidlink}

\begin{document}

\title{ABounD: Adversarial Boundary-Driven Few-Shot Learning for Multi-Class Anomaly Detection} 


\author{Runzhi Deng\inst{1} \and
Yundi Hu\inst{1} \and
Xinshuang Zhang\inst{1} \and
Zhao Wang\inst{2} \and
Xixi Liu\inst{2} \and
Wang-Zhou Dai\inst{1} \and
Caifeng Shan\inst{1} \and
Fang Zhao\inst{1\dag}}

\authorrunning{R.~Deng et al.}

\institute{School of Intelligence Science and Technology, Nanjing University, Suzhou, Jiangsu, China \and
China Mobile Zijin Innovation Institute, Nanjing, Jiangsu, China \\
\email{\{rzdeng, huyd\}@smail.nju.edu.cn}}

\maketitle

\begin{abstract}
Few-shot multi-class industrial anomaly detection identifies diverse defects across multiple categories using a single unified model and limited normal samples. Although vision-language models offer strong generalization, modeling multiple distinct category manifolds concurrently without actual anomalous data causes feature space collapse and cross-class interference. Consequently, existing methods often fail to balance scalability and precision, leading to either isolated single-class retraining or excessively loose decision margins. To address this limitation, we present a one-for-all learning framework called ABounD that unites semantic concept anchoring with geometric boundary optimization. This method employs two lightweight mechanisms to resolve multi-class ambiguity. First, the Dynamic Concept Fusion module generates class-adaptive semantic anchors via query-aware hierarchical calibration, disentangling overlapping category concepts. Second, using these anchors, the Adversarial Boundary Forging module constructs a tight, class-tailored decision margin by synthesizing adversarial boundary-level fence features to prevent cross-class boundary blurring. Optimized in a single stage, ABounD removes the requirement for isolated per-category retraining in few-shot settings. Experiments on seven industrial benchmarks show that the proposed method achieves state-of-the-art detection and localization performance for multi-class few-shot anomaly detection while maintaining low computational costs during training and inference.
\end{abstract}

\section{Introduction}

\begin{figure*}[t]
    \centering
    \includegraphics[width=\linewidth]{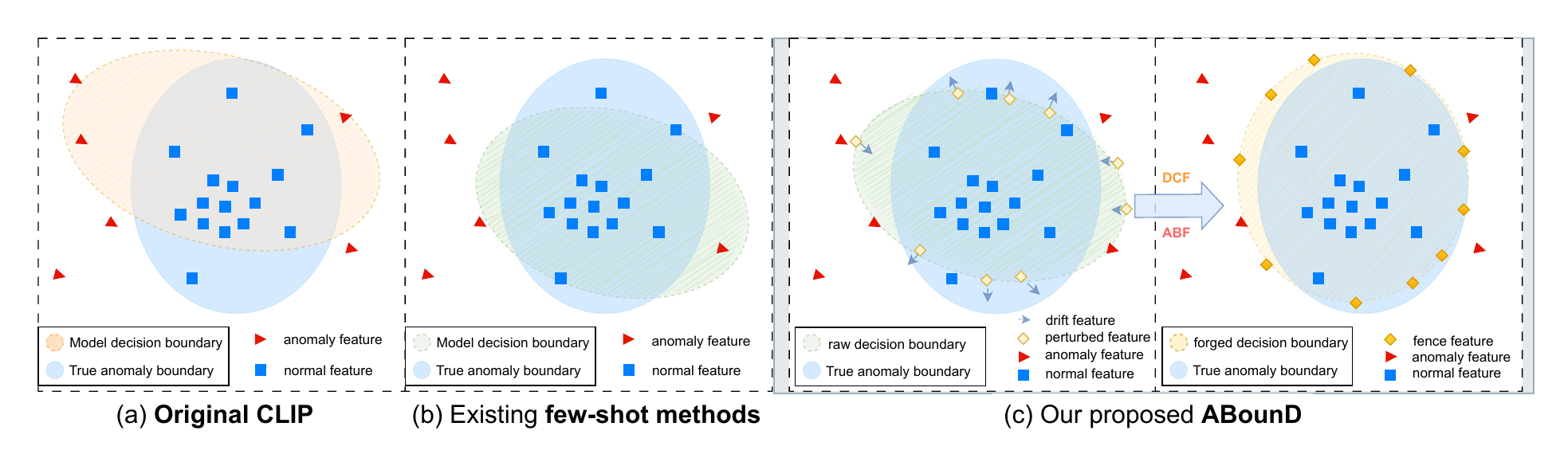}
    \vspace{-10pt}
    \caption{\small Decision boundary intuition for few-shot multi-class anomaly detection. (a) Original CLIP~\cite{radford2021learning} forms a loose boundary that overlaps anomalies, causing false negatives. (b) Existing few-shot methods~\cite{li2024promptad, lv2025one} struggle in the unified multi-class few-shot setting. They suffer from feature space collapse and cross-class interference, resulting in either overly tight boundaries that reject atypical normal samples or vague margins that fail to isolate fine-grained anomalies. (c) ABounD jointly learns Dynamic Concept Fusion and Adversarial Boundary Forging. The former produces class-adaptive semantic anchors, while the latter constructs fence features via projected gradient descent\cite{mkadry2017towards} to shape a sharp decision boundary for multiple classes concurrently. The framework is optimized in a single stage under a concept-boundary loss.}
    \label{fig:motivation}
    \vspace{-8pt}
\end{figure*}

Visual anomaly detection is essential for industrial quality control. A challenging setting is few-shot multi-class anomaly detection \cite{yan2024anomalysd, you2022unified, lu2023hierarchical, he2024diffusion}, where the objective is to identify defects across multiple product categories using only a few normal samples per class. This setting reflects real-world manufacturing scenarios, where anomalies are infrequent and collecting extensive normal data for every new category is cost-prohibitive.

Vision-language models, including CLIP and recent multimodal large language models, align visual features with natural language concepts \cite{radford2021learning, hurst2024gpt, yang2025qwen3, brown2020language, li2023blip}. Benefiting from large-scale pre-training, recent zero-shot approaches show that textual priors can achieve universal anomaly detection without target domain training \cite{jeong2023winclip, zhou2023anomalyclip, cao2024adaclip, gao2025adaptclip, aota2023zero, wang2025zero, deng2025vmad}. Concurrently, to bridge the modality gap, few-shot approaches incorporate limited normal samples to guide the unified detection process \cite{lv2025one, zhu2024toward}. Furthermore, a rapidly growing number of recent studies investigate complex industrial reasoning using multi-stage logic and region-of-interest tokenization \cite{li2025iad, chao2025anomalyr1, miao2025agentiad, li2025triad, zhang2025eiad, zhao2025omniad, guan2025emit, liao2025ad, mokhtar2025detect, zeng2025lr, chen2025can, xu2025towards, jiang2024mmad}.

Despite these advancements, a geometric challenge remains. In the multi-class few-shot setting without actual abnormal training data, the learned representations must adapt to intra-class variations while remaining discriminative against subtle anomalies. Previous methods often fail to balance this trade-off. First, to achieve high precision with limited normal samples, few-shot and unsupervised approaches often revert to a single-class paradigm \cite{li2024promptad, fang2023fastrecon, roth2022towards, gudovskiy2022cflow, liu2023simplenet, zavrtanik2021draem, wyatt2022anoddpm}. These models require isolated retraining for every new object category, which limits multi-class scalability. Second, zero-shot universal methods maintain unified architectures but rely on passive semantic alignment to establish generic normality \cite{zhou2023anomalyclip, gao2025adaptclip, cao2024adaclip, wang2025zero}. In a multi-class feature space, passive alignment without explicit geometric boundary optimization causes cross-class interference. This ambiguity leads to either excessively loose margins that misclassify atypical normal samples or vague frontiers that fail to detect fine-grained anomalies.

To address this bottleneck, we propose ABounD, an adversarial boundary-driven few-shot learning framework. ABounD tightly couples dynamic semantic anchoring with explicit decision-boundary optimization. The core insight is that a precise multi-class boundary requires a stable class-specific anchor, and a semantic anchor requires a geometric margin defended against anomalies. 

Specifically, the Dynamic Concept Fusion module generates adaptive prompts by combining generalizable mixture-of-experts semantics \cite{jacobs1991adaptive, jordan1994hierarchical} with learnable class-adaptive cues conditioned on the global features of the image. This establishes a semantic anchor for each normal class. Using this anchor, the Adversarial Boundary Forging module constructs a class-aware decision margin by generating boundary-level fence features via projected gradient descent perturbations \cite{mkadry2017towards, deng2026generative, ngo2019fence}. Guided by balance and dispersion objectives, the module synthesizes adversarial hard examples that approximate the semantic frontier between normal and abnormal states, which encourages a compact boundary around the normal manifold. Unlike decoupled or stage-wise pipelines \cite{ma2025aa, zuo2024clip, wang2025robust}, these two mechanisms operate in a single stage under a concept-boundary loss, where the dynamic anchor guides adversarial generation and the constructed boundary refines the semantic representation.

In summary, this work introduces ABounD, a single-stage learning framework that addresses feature collapse and boundary ambiguity in few-shot multi-class anomaly detection, moving beyond the restrictive single-class paradigm. Furthermore, it proposes a mutually reinforcing mechanism, where the Dynamic Concept Fusion module provides class-adaptive semantic anchors, and the Adversarial Boundary Forging module constructs a discriminative geometric boundary using synthesized hard examples. Finally, the proposed method achieves state-of-the-art detection and localization performance on the MVTec-AD and VisA benchmarks, demonstrating scalability and robustness without requiring actual anomalous training data.
\section{Related Work}

\subsection{Unsupervised and Multi-Class Anomaly Detection}
Traditional unsupervised anomaly detection relies on abundant normal samples to identify defective patterns. These methods primarily fall into reconstruction-based paradigms, which highlight anomalies via large pixel- or feature-level reconstruction errors \cite{bergmann2019mvtec, matthias2007weakly, jezek2021deep, wyatt2022anoddpm}, and embedding-based approaches, which measure deviations from a modeled normal feature manifold \cite{roth2022towards, zou2022spot, defard2021padim, mishra2021vt, deng2009imagenet, gudovskiy2022cflow, liu2023simplenet, zavrtanik2021draem}. Despite achieving strong performance on individual categories, these conventional paradigms are severely bottlenecked by a rigid one-model-per-class constraint. They demand isolated training phases and extensive normal datasets for each distinct object class, fundamentally restricting their scalability and rapid deployment in dynamic, multi-category industrial environments. This critical limitation has driven the recent shift towards unified and data-efficient architectures \cite{you2022unified, lu2023hierarchical, he2024diffusion}.

\subsection{VLMs for Zero- and Few-Shot Anomaly Detection}
The advent of vision-language models has introduced a transformative paradigm for data-efficient anomaly detection \cite{radford2021learning, hurst2024gpt, yang2025qwen3, brown2020language, li2023blip}. Leveraging large-scale pre-training, recent pioneering zero-shot methods achieve universal detection via text-driven semantic alignment without any target domain training \cite{jeong2023winclip, zhou2023anomalyclip, cao2024adaclip, gao2025adaptclip, aota2023zero, wang2025zero, deng2025vmad}. To further bridge the modality gap under limited supervision, few-shot approaches incorporate lightweight adapters, in-context learning, or specialized prompt tuning to contrast visual features against learnable normal and abnormal text embeddings \cite{li2024promptad, zhu2024toward, lv2025one, zhao2023omnial, fang2023fastrecon}. Concurrently, an explosive growth of research explores empowering Multimodal Large Language Models for complex, multi-stage industrial reasoning and region-of-interest tokenization \cite{li2025iad, chao2025anomalyr1, miao2025agentiad, li2025triad, zhang2025eiad, zhao2025omniad, guan2025emit, liao2025ad, mokhtar2025detect, zeng2025lr, chen2025can, xu2025towards, jiang2024mmad}.

Despite these remarkable advancements, existing few-shot methods face critical limitations when applied to unified multi-class scenarios containing exclusively normal samples. To maintain high localization precision, leading few-shot techniques often regress to the target-specific single-class paradigm, thereby sacrificing universal scalability. Furthermore, current multi-class prompt-learning strategies predominantly focus on optimizing overarching semantic concepts but critically fail to explicitly sculpt the geometric decision boundary. This passive semantic alignment inevitably leads to cross-class interference in a crowded feature space, resulting in either excessively loose margins that misclassify atypical normal variations or vague frontiers that entirely overlook subtle and fine-grained anomalies. Distinct from these decoupled pipelines, our proposed ABounD framework uniquely synthesizes dynamic concept anchoring with adversarial boundary forging, explicitly crafting a sharp, highly discriminative decision boundary within a unified single-stage learning paradigm.
\section{Methodology}
\label{sec:methodology}

\begin{figure*}[t]
\centering
\includegraphics[width=\linewidth]{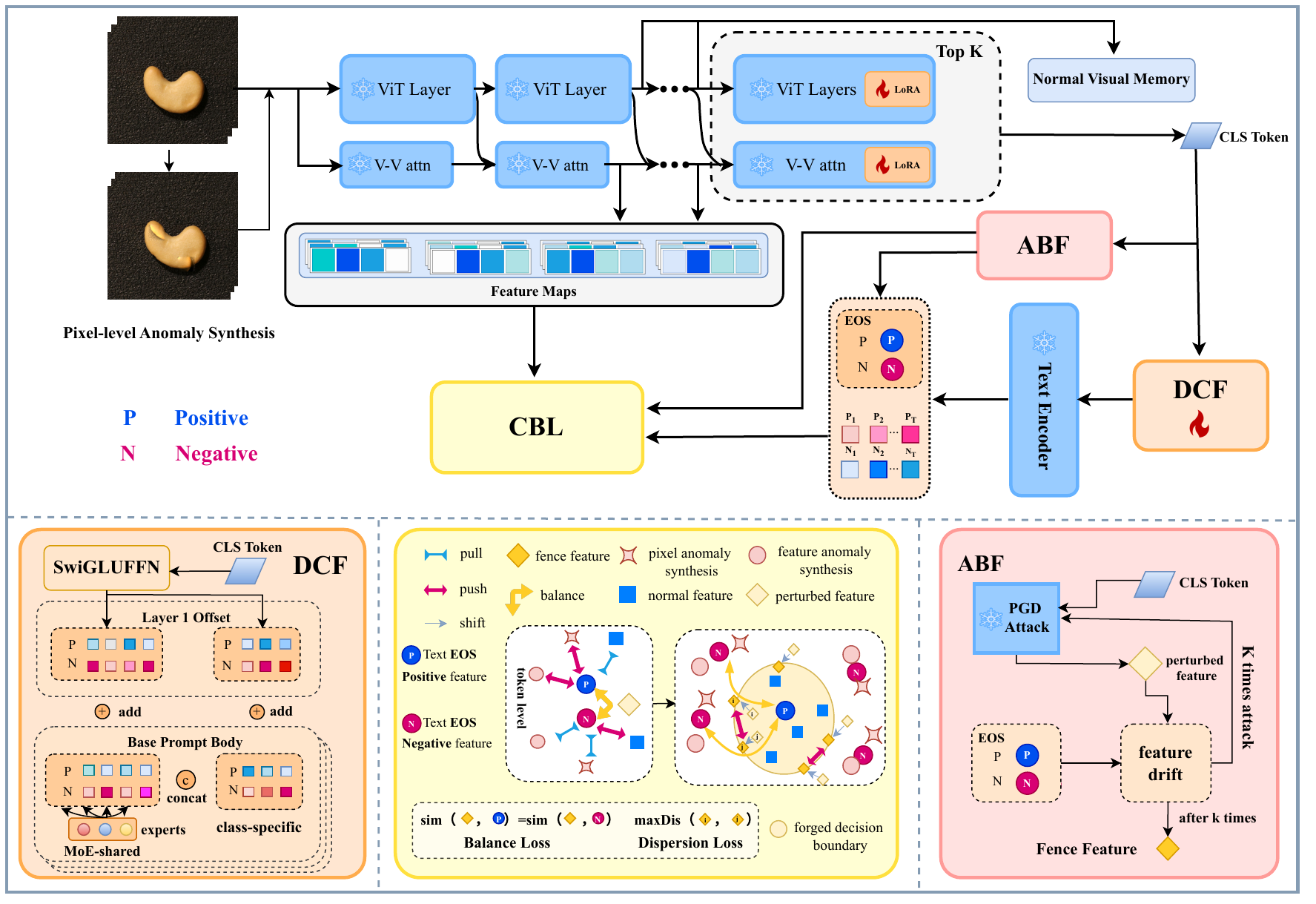} 
\vspace{-12pt}
\caption{
Overview of the \textbf{ABounD} framework, which introduces two core components: \textbf{Dynamic Concept Fusion (DCF)} and \textbf{Adversarial Boundary Forging (ABF)}. 
DCF generates class-adaptive prompts by combining generalizable Mixture-of-Experts (MoE) semantics with learnable class-adaptive cues based on the global visual representation. 
ABF guides feature learning by performing PGD-based adversarial perturbations near the decision boundary, where balance and dispersion objectives ensure well-distributed hard examples. 
The framework is optimized in a single stage under the \textit{Concept-Boundary Loss (CBL)}, enabling robust few-shot multi-class anomaly detection.
}
\label{fig2}
\end{figure*}

\subsection{Preliminaries and Problem Setting}
This work addresses few-shot multi-class anomaly detection under a unified learning paradigm. 
Let the training set be defined as $\mathcal{D}_{train} = \{(\mathbf{x}_i, y_i) \mid y_i \in \mathcal{C}_{seen}\}$, consisting entirely of normal samples from seen classes. 
In the $K$-shot setting, only $K$ normal support images are available per class. 
The objective is to learn an anomaly scoring function $\mathcal{A}(\mathbf{x}): \mathcal{X} \rightarrow [0, 1]$ that generalizes to unseen samples and distinguishes anomalies without accessing actual anomalous data during training. 
This task is challenging because modeling multiple category manifolds concurrently with severely limited normal data causes feature space collapse and semantic ambiguity between classes. An overview of the proposed ABounD framework, designed to address these challenges, is illustrated in Fig.~\ref{fig2}.

\subsection{Dynamic Concept Fusion: The Semantic Anchor}
To establish the semantic anchor for the proposed mutually reinforcing mechanism, this study introduces \textbf{Dynamic Concept Fusion (DCF)}. 
Existing static prompt learning methods \cite{zhou2022learning} combine category-specific attributes with shared anomaly patterns, which restricts cross-domain flexibility and increases feature interference in multi-class scenarios. 
The proposed structural design decomposes transferable anomaly patterns from category-conditioned manifolds and hierarchically constructs dynamic contextual calibrations. 
This disentanglement enables the formation of a structured and adaptive semantic anchor for subsequent boundary optimization.

\subsubsection{Universal-Specific Decomposition via MoE}
Instead of relying on monolithic representations, a dual-pathway mechanism is employed. 
The first pathway captures structural anomaly patterns applicable across different categories using a Mixture-of-Experts (MoE) module with $N$ learnable bases. 
A gating network assigns weights based on the global visual token $\mathbf{v}_{cls} \in \mathbb{R}^D$ to construct a shared embedding $\mathbf{P}_{shared}$:
\begin{equation}
    \mathbf{P}_{shared}(\mathbf{v}_{cls}) = \sum_{k=1}^N \alpha_k \cdot \mathbf{B}_k,
\end{equation}
where $\mathbf{B}_k \in \mathbb{R}^{L \times D}$ represents the $k$-th basis and $\alpha_k$ is the routing weight computed as $\alpha_k = \text{Softmax}(\mathbf{v}_{cls}\mathbf{W}_{gate})_k$, with $\mathbf{W}_{gate} \in \mathbb{R}^{D \times N}$ representing a learnable projection matrix. 
Concurrently, the second pathway retrieves a learnable parameter bank $\mathbf{P}_{spec}^{c} \in \mathbb{R}^{L \times D}$ to capture the semantic manifold of category $c$. 
The foundational static prompt is formed by concatenating these embeddings along the sequence dimension.

\subsubsection{Query-Aware Semantic Calibration}
Although the decomposed prompt provides a category-level prior, it often fails to align with specific visual queries due to intra-class variations, such as lighting or texture shifts. To address this limitation, a query-adaptive residual offset is introduced. Given the visual query $\mathbf{v}_{cls}$, a SwiGLU-based gating network filters irrelevant visual noise to predict a modulation vector $\mathbf{P}_{mod} \in \mathbb{R}^D$:
\begin{equation}
    \mathbf{P}_{mod}(\mathbf{v}_{cls}) = \text{SwiGLU}(\mathbf{v}_{cls}) = \left( \text{SiLU}(\mathbf{v}_{cls}\mathbf{W}_g) \odot (\mathbf{v}_{cls}\mathbf{W}_u) \right)\mathbf{W}_d.
\end{equation}
In this formulation, $\mathbf{W}_g, \mathbf{W}_u \in \mathbb{R}^{D \times D_{mid}}$ and $\mathbf{W}_d \in \mathbb{R}^{D_{mid} \times D}$ are learnable weight matrices, and $\odot$ represents the element-wise product. This dynamic calibration ensures that the textual space adapts to the visual context of the query image.

\subsubsection{Hierarchical Context Assembly}
To incorporate these adaptive concepts without affecting the representation capacity of the pre-trained vision-language model, a hierarchical context assembly strategy is implemented.

\paragraph{Shallow Assembly.}
For the initial input layer, the input sequence is constructed by combining the static prompt and the query-aware calibration. The final input textual sequence $\mathbf{E}_{input}$ is formulated as:
\begin{equation}
    \mathbf{E}_{input} = \big[ \ \big(\mathbf{P}_{shared}(\mathbf{v}_{cls}) \parallel \mathbf{P}_{spec}^{c}\big) \oplus \mathbf{P}_{mod}(\mathbf{v}_{cls}) \ \big],
\end{equation}
where $\parallel$ denotes sequence-level concatenation resulting in a length of $2L$, and $\oplus$ denotes element-wise addition where the query-adaptive vector $\mathbf{P}_{mod} \in \mathbb{R}^D$ is broadcasted across all $2L$ tokens. This formulation encodes both the universal and category-adaptive contexts while calibrating them with the visual query.

\paragraph{Deep Context Refreshing.}
In deeper transformer layers denoted by index $l$, context refreshing is introduced to prevent semantic dilution. Layer-specific prompts $\mathbf{P}^{(l)}$ are injected to replace the keys and values at corresponding positions, using only the static components:
\begin{equation}
    \mathbf{P}^{(l)} = \big[ \ \mathbf{P}_{shared}^{(l)}(\mathbf{v}_{cls}) \ \parallel \ \mathbf{P}_{spec}^{c,(l)} \ \big].
\end{equation}
Omitting the query-aware calibration $\mathbf{P}_{mod}$ in intermediate layers prevents the model from overfitting to low-level visual statistics, thereby preserving stable semantic anchors throughout the network depth.

\subsection{Adversarial Boundary Forging: The Geometric Sculptor}
Relying solely on semantic anchors leaves the decision boundaries between normal variations and anomalies ambiguous. 
To address this ambiguity, \textbf{Adversarial Boundary Forging (ABF)} is proposed. 
Inspired by generative feature drifting \cite{deng2026generative}, boundary identification is formulated as a dynamic process where normal features evolve towards a class-tailored fence region~\cite{ngo2019fence} characterized by maximum predictive uncertainty.

\paragraph{Phase 1: Boundary Evolution via Drifting Field.}
An efficient feature-level drifting field is defined to drive normal latent features towards the semantic frontier of their respective classes. 
First, a \textit{Balance Potential} is introduced to locate the boundary geometrically. 
Let $\mathbf{p}_{pos}$ and $\mathbf{p}_{neg}$ be the $L_2$-normalized text embeddings obtained by feeding the respective class-adaptive textual sequence $\mathbf{E}_{input}$ into the text encoder. 
This term encourages the latent feature $\mathbf{v}$ to be semantically equidistant from the normal and abnormal textual anchors:
\begin{equation}
    J_{\text{bal}}(\mathbf{v}) = \left| \text{sim}(\mathbf{v}, \mathbf{p}_{pos}) - \text{sim}(\mathbf{v}, \mathbf{p}_{neg}) \right|,
\end{equation}
where $\text{sim}(\cdot, \cdot)$ denotes the cosine similarity function.

Second, to ensure broad coverage of the boundary and prevent collapse into local minima, a \textit{Diversity Potential} is introduced. This potential measures the average pairwise distance within a batch of features $V = \{\mathbf{v}_1, \dots, \mathbf{v}_B\}$ of size $B$ to ensure well-distributed hard examples:
\begin{equation}
    J_{\text{div}}(V) = \frac{2}{B(B-1)} \sum_{i<j} \| \mathbf{v}_{i} - \mathbf{v}_{j} \|_2,
\end{equation}
where $\mathbf{v}_i$ and $\mathbf{v}_j$ represent individual feature vectors within the batch.

The dynamic evolution minimizes the total \textit{Drifting Potential} $J_{\text{drift}}$, optimizing both boundary alignment and feature dispersion across the batch:
\begin{equation}
    J_{\text{drift}}(V) = \frac{1}{B} \sum_{i=1}^B J_{\text{bal}}(\mathbf{v}_i) - \beta J_{\text{div}}(V),
\end{equation}
where $\beta$ is a balancing hyper-parameter. 
This objective is optimized iteratively via $T$-step projected gradient descent within the latent space, yielding the final boundary features $\mathbf{v}_{fence}$.

\paragraph{Phase 2: Boundary Sculpting via Stop-Gradient.}
Once the boundaries are populated with adversarial perturbations $\boldsymbol{\delta}$, 
the drifted features are constructed as $\mathbf{v}_{fence} = \mathbf{v} + \text{sg}(\boldsymbol{\delta})$, 
where the stop-gradient operator $\text{sg}(\cdot)$ is applied during the projected gradient descent update to avoid back-propagating through the iterative perturbation trajectory. 
Gradients from the entropy objective are allowed to flow back to the visual encoder and prompt parameters via $\mathbf{v}_{fence}$, 
ensuring that boundary optimization directly updates the model representation while keeping the adversarial search stable:
\begin{equation}
    \mathcal{L}_{\text{ABF}} = \sum_{s \in \{\text{pos}, \text{neg}\}} P(s \mid \mathbf{v}_{fence}) \log P(s \mid \mathbf{v}_{fence}),
\end{equation}
where the probability $P(s \mid \mathbf{v})$ is computed via a softmax function over the cosine similarities scaled by a temperature parameter $\tau$:
\begin{equation}
    P(s \mid \mathbf{v}) = \frac{\exp(\text{sim}(\mathbf{v}, \mathbf{p}_s)/\tau)}{\sum_{s' \in \{\text{pos}, \text{neg}\}} \exp(\text{sim}(\mathbf{v}, \mathbf{p}_{s'})/\tau)}.
\end{equation}
Here, $s$ denotes the binary semantic state (\ie, positive or negative normality), $s'$ serves as the summation index over these two states, and $\mathbf{p}_s$ represents the dynamic textual anchor ($\mathbf{p}_{pos}$ for normal, $\mathbf{p}_{neg}$ mapped from learnable defect templates). 
In a multi-class setting, this objective constructs the decision boundary around each specific normal manifold, preventing the misclassification of out-of-distribution defects as normal variations of nearby classes.

\subsection{Training and Inference}
\label{sec:train_infer}

\paragraph{Unified Multi-Class Training.}
The framework updates the DCF module and the visual encoder jointly, keeping the remainder of the vision-language model frozen. To enable spatial supervision without accessing actual anomalous data, pixel-level anomalies are synthesized via a cut-and-paste strategy~\cite{liu2023simplenet} during training, as shown in Fig.~\ref{fig2}. Optimization is governed by the unified \textit{Concept-Boundary Loss}:
\begin{equation}
\mathcal{L}_{\mathrm{CBL}} = \lambda_{\mathrm{abf}} \mathcal{L}_{\mathrm{ABF}} + \lambda_{\mathrm{psg}} \mathcal{L}_{\mathrm{PSG}} + \lambda_{\mathrm{seg}} \mathcal{L}_{\mathrm{SEG}}.
\end{equation}
Here, $\mathcal{L}_{\mathrm{ABF}}$ drives the boundary optimization, stabilized by two regularization terms: the prompt semantic grounding loss ($\mathcal{L}_{\mathrm{PSG}}$)~\cite{lv2025one}, which ensures DCF prompts align semantically with normal visual patches, and the segmentation loss ($\mathcal{L}_{\mathrm{SEG}}$), which supervises the synthetic anomalies. Detailed formulations are provided in the supplementary material.

\paragraph{Zero-Prior Multi-Class Inference.} 
During inference, the framework operates in an automated manner without requiring prior knowledge of the test image category, enabling practical deployment in label-free multi-class environments.

\textbf{Automatic Class Routing.}
For each class $k \in \mathcal{C}_{seen}$, the normal manifold is represented using the $L_2$-normalized mean $\boldsymbol{\mu}_k$ of the global visual features extracted from its few-shot support samples. 
Given a test image without a label prior, it is automatically routed to its most probable category $k_m$ by identifying the class center that yields the maximum cosine similarity with the global feature of the test image.

\textbf{Memory-Fused Anomaly Score.}
Once routed to class $k_m$, the final spatial anomaly map is computed as a fusion of text-driven and visual-driven metrics. 
The text map $\mathbf{M}_{\text{text}}$ computes the anomaly response using $1 - \text{sim}(\mathbf{v}_{\text{patch}}, \mathbf{p}_{\text{pos}})$ between the visual patch features $\mathbf{v}_{\text{patch}}$ and the positive query-calibrated prompt $\mathbf{p}_{\text{pos}}$. 
The visual map $\mathbf{M}_{\text{vis}}$ is derived from the nearest-neighbor Euclidean distance within the class-aligned normal memory bank. 
The final spatial map and the image-level anomaly score $S$ are obtained as:
\begin{equation} 
\mathbf{M} = \mathbf{M}_{\text{text}} + \gamma \mathbf{M}_{\text{vis}}, \quad S = \max(\mathbf{M}).
\end{equation}
where $\gamma$ acts as a balancing scalar.
\section{Experiments}
\begin{table*}[t]
\centering
\caption{Comparison of image-level anomaly classification using the $\text{AUROC}_I$ metric across seven industrial domains. The best results are highlighted in bold. All few-shot baselines are re-evaluated under the same unified multi-class setting for fair comparison(our ViT-L/14 adaptation of PromptAD outperforms the one in~\cite{lv2025one}, see Appendix). For IIPAD, the results on MVTec and VisA are cited from the original publication, whereas the remaining results are based on local reproduction. The results for InCTRL are cited directly, with `-' indicating unevaluated datasets.}
\label{tab:main_image}
\resizebox{\textwidth}{!}{
\begin{tabular}{cc|ccccccc|c}
\toprule
\multirow{2}{*}{\textbf{Shots}} & \multirow{2}{*}{\textbf{Methods}} & \multicolumn{8}{c}{\textbf{Industrial Benchmarks}} \\
\cmidrule(lr){3-10}
 & & MVTec & VisA & BTAD & MVTec3D & DTD & MPDD & Real-IAD & AVG \\
\midrule
\multirow{3}{*}{\textbf{0}} 
 & WinCLIP~\cite{jeong2023winclip}        & 90.4 & 75.5 & 68.2 & 69.4 & 95.1 & 61.5 & 67.0 & 75.3 \\
 & AnomalyCLIP~\cite{zhou2023anomalyclip} & 91.6 & 82.0 & 88.3 & 73.9 & 93.9 & 77.5 & 69.5 & 82.4 \\
 & AdaptCLIP~\cite{gao2025adaptclip}      & 93.5 & 84.8 & 91.0 & 78.6 & 96.0 & 73.6 & 74.2 & 84.5 \\
\midrule
\multirow{3}{*}{\textbf{1}} 
 & PromptAD~\cite{li2024promptad}        & 90.9$\pm$0.4 & 85.9$\pm$0.7 & 88.5$\pm$0.6 & 70.5$\pm$0.8 & 91.3$\pm$0.4 & 68.3$\pm$1.1 & 72.6$\pm$0.7 & 81.1 \\
 & IIPAD~\cite{lv2025one}                    & 94.2 & 85.4 & 89.6$\pm$0.5 & 77.9$\pm$0.9 & 98.3$\pm$0.2 & 75.1$\pm$1.4 & 81.6$\pm$0.3 & 86.0 \\
 & \textbf{ABounD (Ours)}                 & \textbf{94.8$\pm$0.1} & \textbf{87.3$\pm$0.3} & \textbf{93.3$\pm$0.2} & \textbf{80.7$\pm$0.4} & \textbf{98.9$\pm$0.1} & \textbf{82.8$\pm$0.4} & \textbf{84.3$\pm$0.5} & \textbf{88.9} \\
\midrule
\multirow{4}{*}{\textbf{2}} 
 & PromptAD~\cite{li2024promptad}        & 91.5$\pm$0.3 & 86.7$\pm$0.7 & 87.3$\pm$0.5 & 72.8$\pm$0.7 & 92.1$\pm$0.3 & 69.1$\pm$1.2 & 74.5$\pm$0.6 & 82.0 \\
 & InCTRL~\cite{zhu2024toward}            & 94.0$\pm$1.5 & 85.8$\pm$2.2 & - & - & - & - & - & - \\
 & IIPAD~\cite{lv2025one}                    & 95.7 & 86.7 & 91.3$\pm$0.4 & 79.1$\pm$0.8 & 98.7$\pm$0.2 & 74.7$\pm$1.2 & 82.3$\pm$0.4 & 86.9 \\
 & \textbf{ABounD (Ours)}                 & \textbf{96.8$\pm$0.1} & \textbf{89.4$\pm$0.3} & \textbf{93.7$\pm$0.2} & \textbf{81.5$\pm$0.3} & \textbf{99.1$\pm$0.1} & \textbf{83.2$\pm$0.3} & \textbf{86.7$\pm$0.4} & \textbf{90.1} \\
\midrule
\multirow{4}{*}{\textbf{4}} 
 & PromptAD~\cite{li2024promptad}        & 93.4$\pm$0.3 & 86.9$\pm$0.9 & 87.9$\pm$0.4 & 74.5$\pm$0.6 & 92.6$\pm$0.3 & 71.8$\pm$0.9 & 76.3$\pm$0.5 & 83.3 \\
 & InCTRL~\cite{zhu2024toward}            & 94.5$\pm$1.8 & 87.7$\pm$1.9 & - & - & - & - & - & - \\
 & IIPAD~\cite{lv2025one}                    & 96.1 & 88.3 & 91.2$\pm$0.5 & 80.9$\pm$0.7 & 98.7$\pm$0.2 & 76.4$\pm$1.5 & 84.2$\pm$0.5 & 88.0 \\
 & \textbf{ABounD (Ours)}                 & \textbf{97.0$\pm$0.1} & \textbf{90.3$\pm$0.8} & \textbf{93.5$\pm$0.1} & \textbf{83.1$\pm$0.5} & \textbf{99.1$\pm$0.1} & \textbf{84.7$\pm$0.4} & \textbf{87.6$\pm$0.3} & \textbf{90.8} \\
\bottomrule
\end{tabular}
}
\end{table*}

\begin{table*}[t]
\centering
\caption{Comparison of pixel-level anomaly segmentation using the $\text{AUPRO}_P$ metric across seven industrial domains. Consistent with Table 1, all few-shot baselines are evaluated under the same unified multi-class setting. For IIPAD, the results on MVTec and VisA are cited from the original publication, whereas the remaining results are reproduced locally.}
\label{tab:main_pixel_pro}
\resizebox{\textwidth}{!}{
\begin{tabular}{cc|ccccccc|c}
\toprule
\multirow{2}{*}{\textbf{Shots}} & \multirow{2}{*}{\textbf{Methods}} & \multicolumn{8}{c}{\textbf{Industrial Benchmarks}} \\
\cmidrule(lr){3-10}
 & & MVTec & VisA & BTAD & MVTec3D & DTD & MPDD & Real-IAD & AVG \\
\midrule
\multirow{3}{*}{\textbf{0}} 
 & WinCLIP~\cite{jeong2023winclip}        & 64.6 & 56.8 & 27.3 & 54.2 & 57.8 & 48.9 & 56.5 & 52.3 \\
 & AnomalyCLIP~\cite{zhou2023anomalyclip} & 81.4 & 87.0 & 74.8 & 76.8 & 92.3 & 88.7 & 78.4 & 82.8 \\
 & AdaptCLIP~\cite{gao2025adaptclip}      & 83.1 & 88.7 & 77.4 & 81.5 & 94.6 & 90.2 & 82.3 & 85.4 \\
\midrule
\multirow{3}{*}{\textbf{1}} 
 & PromptAD~\cite{li2024promptad}        & 87.3$\pm$0.2 & 82.2$\pm$0.9 & 72.5$\pm$0.4 & 87.8$\pm$0.6 & 93.1$\pm$0.3 & 86.4$\pm$0.8 & 79.5$\pm$0.5 & 84.1 \\
 & IIPAD~\cite{lv2025one}                    & 89.8 & 87.3 & 72.6$\pm$0.3 & 91.0$\pm$0.4 & 94.7$\pm$0.2 & 90.9$\pm$0.3 & 85.4$\pm$0.4 & 87.4 \\
 & \textbf{ABounD (Ours)}                 & \textbf{91.1$\pm$0.1} & \textbf{88.6$\pm$0.2} & \textbf{79.0$\pm$0.2} & \textbf{93.0$\pm$0.1} & \textbf{96.1$\pm$0.1} & \textbf{92.9$\pm$0.1} & \textbf{91.1$\pm$0.1} & \textbf{90.3} \\
\midrule
\multirow{3}{*}{\textbf{2}} 
 & PromptAD~\cite{li2024promptad}        & 88.0$\pm$0.2 & 81.6$\pm$0.6 & 73.1$\pm$0.5 & 88.2$\pm$0.5 & 93.8$\pm$0.4 & 87.0$\pm$0.7 & 80.1$\pm$0.4 & 84.5 \\
 & IIPAD~\cite{lv2025one}                    & 90.3 & 87.9 & 72.7$\pm$0.3 & 90.6$\pm$0.4 & 95.3$\pm$0.2 & 90.8$\pm$0.2 & 85.7$\pm$0.3 & 87.6 \\
 & \textbf{ABounD (Ours)}                 & \textbf{91.9$\pm$0.1} & \textbf{88.9$\pm$0.2} & \textbf{79.3$\pm$0.2} & \textbf{93.1$\pm$0.1} & \textbf{96.2$\pm$0.1} & \textbf{93.4$\pm$0.1} & \textbf{92.1$\pm$0.1} & \textbf{90.7} \\
\midrule
\multirow{3}{*}{\textbf{4}} 
 & PromptAD~\cite{li2024promptad}        & 88.4$\pm$0.2 & 82.0$\pm$0.8 & 74.0$\pm$0.4 & 88.9$\pm$0.3 & 94.4$\pm$0.2 & 87.6$\pm$0.6 & 80.9$\pm$0.5 & 85.2 \\
 & IIPAD~\cite{lv2025one}                    & 91.2 & 88.3 & 72.6$\pm$0.2 & 91.0$\pm$0.3 & 95.5$\pm$0.2 & 90.7$\pm$0.3 & 87.2$\pm$0.4 & 88.1 \\
 & \textbf{ABounD (Ours)}                 & \textbf{92.3$\pm$0.1} & \textbf{89.2$\pm$0.3} & \textbf{80.4$\pm$0.2} & \textbf{93.8$\pm$0.1} & \textbf{96.3$\pm$0.1} & \textbf{93.5$\pm$0.1} & \textbf{92.6$\pm$0.1} & \textbf{91.2} \\
\bottomrule
\end{tabular}
}
\end{table*}

\subsection{Experimental Setup}
\paragraph{Datasets and Metrics.} 
The proposed method is evaluated on seven industrial benchmarks: MVTec-AD~\cite{bergmann2019mvtec}, VisA~\cite{zou2022spot}, BTAD~\cite{mishra2021vt}, MVTec-3D~\cite{bergmann2021mvtec}, DTD~\cite{aota2023zero}, MPDD~\cite{jezek2021deep}, and Real-IAD~\cite{wang2024real}. Following the unified multi-class few-shot protocol, a single model is jointly trained and evaluated across all categories using only $K$ normal images per class. Detection and localization performance are primarily measured via the image-level Area Under the Receiver Operating Characteristic Curve ($\text{AUROC}_I$) and pixel-level Per-Region Overlap ($\text{PRO}$). Additional metrics and per-category breakdowns are provided in the supplementary material.

\paragraph{Implementation Details and Baselines.}
All few-shot baselines are evaluated under the unified one-for-all protocol using the officially released implementations and public checkpoints. For IIPAD, the original publication reports results on MVTec-AD and VisA under the same protocol; these numbers are cited directly, and the results for the remaining datasets are reproduced under the proposed setup. For WinCLIP, AnomalyCLIP, and AdaptCLIP, the results are reported directly from the original publications. For AdaptCLIP, the results for the AUPRO metric are reproduced locally, as they were not provided in the original publication. The top-1 automatic routing accuracy consistently exceeds 99\% across all benchmarks under the 1-, 2-, and 4-shot settings (details are provided in the supplementary material). Furthermore, using top-2 routing yields nearly identical detection performance, demonstrating robustness to rare routing failures.

\subsection{Main Results}
Tables~\ref{tab:main_image} and \ref{tab:main_pixel_pro} show ABounD consistently achieves top or second-best performance across all benchmarks and shot settings. Unlike IIPAD~\cite{lv2025one}, which uses a pre-trained Q-Former, our framework attains superior accuracy with a lightweight architecture, avoiding heavyweight adapters.

\paragraph{Comparison with Single-Class Paradigms.} 
Table~\ref{tab:one_for_one_comparison} compares our unified one-for-all model against one-for-one methods like PromptAD~\cite{li2024promptad} and WinCLIP~\cite{jeong2023winclip}, which require isolated per-category training. Despite inherent multi-class interference, our framework systematically achieves competitive or superior performance without per-category retraining. This indicates our mutually reinforcing mechanism effectively mitigates cross-class interference.
\begin{table*}[t]
\centering
\caption{Comparison of anomaly detection and localization performance on MVTec-AD and VisA across different few-shot settings. The baseline methods operate under the one-for-one (single-class) paradigm, with results cited from the original publications. In contrast, the proposed method operates under the one-for-all (multi-class) paradigm. The best results are in bold, and the second-best results are underlined.}
\resizebox{\textwidth}{!}{
\begin{tabular}{c l c c c c c c c c}
\toprule
\multirow{3}{*}{Setup} & \multirow{3}{*}{Method} & \multicolumn{4}{c}{MVTec-AD} & \multicolumn{4}{c}{VisA} \\
\cmidrule(lr){3-6} \cmidrule(lr){7-10}
& & \multicolumn{2}{c}{Image-level} & \multicolumn{2}{c}{Pixel-level} & \multicolumn{2}{c}{Image-level} & \multicolumn{2}{c}{Pixel-level} \\
\cmidrule(lr){3-4} \cmidrule(lr){5-6} \cmidrule(lr){7-8} \cmidrule(lr){9-10}
& & AUROC & AUPR & AUROC & PRO & AUROC & AUPR & AUROC & PRO \\

\midrule
\multirow{6}{*}{1-shot} 
& SPADE~\cite{cohen2020sub} & 81.0 & 90.6 & 91.2 & 83.9 & 79.5 & 82.0 & 95.6 & 84.1 \\
& PatchCore~\cite{roth2022towards} & 83.4 & 92.2 & 92.0 & 79.7 & 79.9 & 82.8 & 95.4 & 80.5 \\
& FastRecon~\cite{fang2023fastrecon} & - & - & - & - & - & - & - & - \\
& WinCLIP~\cite{jeong2023winclip} & 93.1 & 96.5 & 95.2 & 87.1 & 83.8 & \underline{85.1} & 96.4 & \underline{85.1} \\
& PromptAD~\cite{li2024promptad} & \underline{94.6} & \underline{97.1} & \underline{95.9} & \underline{87.9} & \underline{86.9} & 88.4 & \underline{96.7} & \underline{85.1} \\
& \textbf{ABounD (One-for-All)} & \textbf{94.8} & \textbf{97.5} & \textbf{96.2} & \textbf{91.1} & \textbf{87.3} & \textbf{89.0} & \textbf{97.4} & \textbf{88.6} \\

\midrule
\multirow{6}{*}{2-shot} 
& SPADE~\cite{cohen2020sub} & 82.9 & 91.7 & 92.0 & 85.7 & 80.7 & 82.3 & 96.2 & 85.7 \\
& PatchCore~\cite{roth2022towards} & 86.3 & 93.8 & 93.3 & 82.3 & 81.6 & 84.8 & 96.1 & 82.6 \\
& FastRecon~\cite{fang2023fastrecon} & 91.0 & - & 95.9 & - & - & - & - & - \\
& WinCLIP~\cite{jeong2023winclip} & 94.4 & 97.0 & 96.0 & 88.4 & 84.6 & 85.8 & 96.8 & \underline{86.2} \\
& PromptAD~\cite{li2024promptad} & \underline{95.7} & \underline{97.9} & \underline{96.2} & \underline{88.5} & \underline{88.3} & \underline{90.0} & \underline{97.1} & 85.8 \\
& \textbf{ABounD (One-for-All)} & \textbf{96.8} & \textbf{98.2} & \textbf{96.6} & \textbf{91.9} & \textbf{89.4} & \textbf{91.8} & \textbf{97.5} & \textbf{88.9} \\

\midrule
\multirow{6}{*}{4-shot} 
& SPADE~\cite{cohen2020sub} & 84.8 & 92.5 & 92.7 & 87.0 & 81.7 & 83.4 & 96.6 & 87.3 \\
& PatchCore~\cite{roth2022towards} & 88.8 & 94.5 & 94.3 & 84.3 & 85.3 & 87.5 & 96.8 & 84.9 \\
& FastRecon~\cite{fang2023fastrecon} & 94.2 & - & \underline{97.0} & - & - & - & - & - \\
& WinCLIP~\cite{jeong2023winclip} & 95.2 & 97.3 & 96.2 & 89.0 & 87.3 & 88.8 & 97.2 & \underline{87.6} \\
& PromptAD~\cite{li2024promptad} & \underline{96.6} & \textbf{98.5} & 96.5 & \underline{90.5} & \underline{89.1} & \underline{90.8} & \underline{97.4} & 86.2 \\
& \textbf{ABounD (One-for-All)} & \textbf{97.0} & \underline{98.3} & \textbf{97.1} & \textbf{92.3} & \textbf{90.3} & \textbf{92.4} & \textbf{97.6} & \textbf{89.2} \\

\bottomrule
\end{tabular}
}

\label{tab:one_for_one_comparison}
\end{table*}

\begin{table}[htbp]
    \centering
    \caption{Comparison with full-shot paradigms. The proposed method remains competitive despite the scarcity of training data.}
    \label{tab:performance_comparison3}
    \small
    \setlength{\tabcolsep}{3pt} 
    \begin{tabular}{l c cc cc}
        \toprule
        \multirow{2.5}{*}{Method} & \multirow{2.5}{*}{Setup} & \multicolumn{2}{c}{MVTec-AD} & \multicolumn{2}{c}{VisA} \\
        \cmidrule(lr){3-4} \cmidrule(lr){5-6}
        & & AUROC$_I$ & AUROC$_P$ & AUROC$_I$ & AUROC$_P$ \\
        \midrule
        \multirow{2}{*}{\textbf{Ours}}
        & 4-shot & 97.0 & 96.9 & 90.3 & 97.6 \\
        & 32-shot & \textbf{97.7} & \textbf{97.5} & \textbf{93.3} & \textbf{98.6} \\
        \midrule
        UniAD~\cite{you2022unified} & Full-shot & 96.5 & 96.8 & 91.9 & 98.6 \\
        OmniAL~\cite{zhao2023omnial} & Full-shot & 97.2 & 98.3 & 87.8 & 96.6 \\
        HVQ-Trans~\cite{lu2023hierarchical} & Full-shot & 98.0 & 97.3 & 93.2 & 98.7 \\
        \bottomrule
    \end{tabular}
\end{table}

As presented in Table~\ref{tab:performance_comparison3}, the proposed framework narrows the performance gap between the few-shot and full-shot paradigms. With only 32 normal samples per class, the method achieves performance comparable to several recent full-shot models, despite the smaller training set.

\begin{figure}[t]
    \centering
    \includegraphics[width=\linewidth]{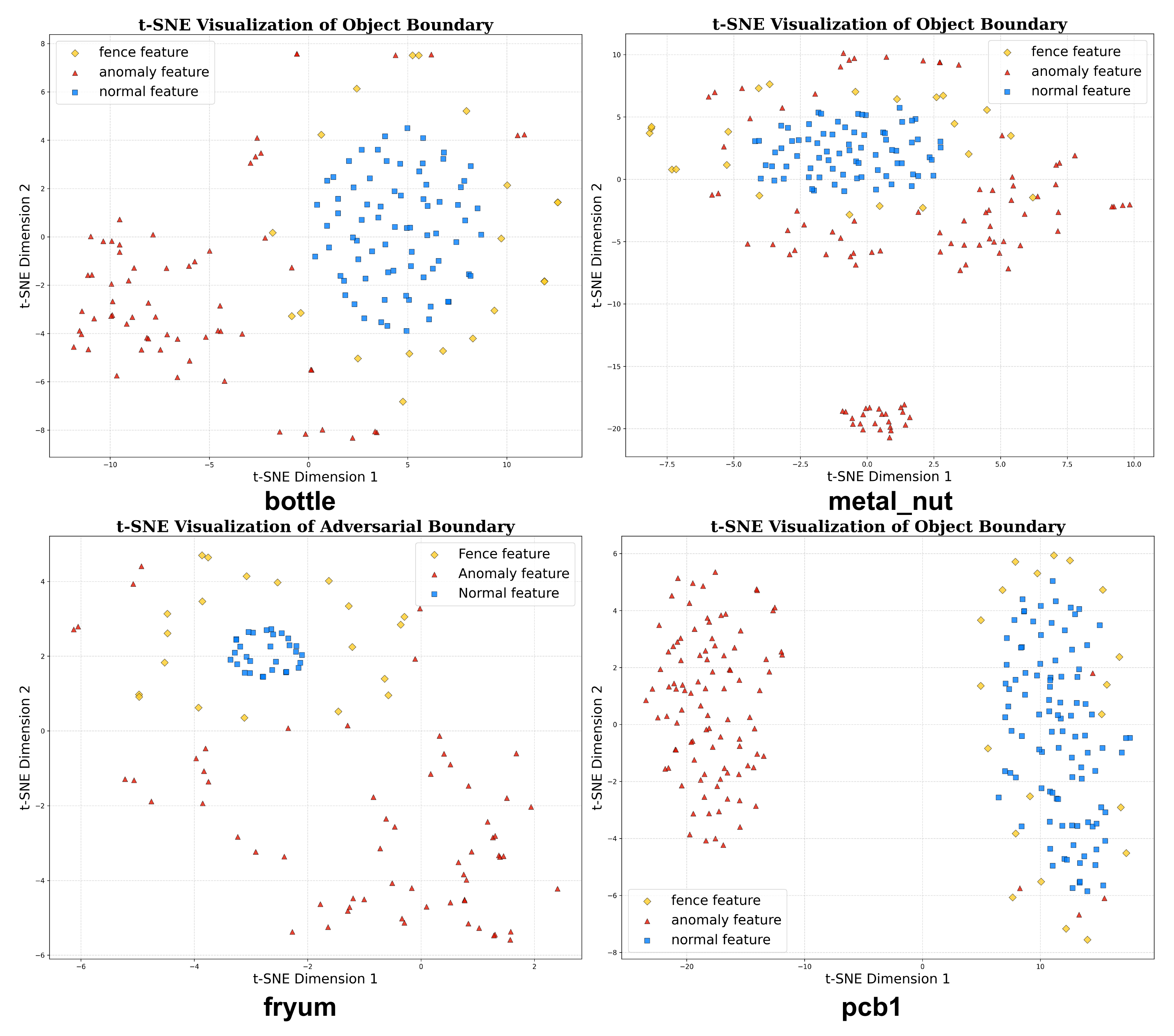}
    \vspace{-15pt}
    \caption{t-SNE visualization of the 1-shot feature space. The synthetic fence features generated by the adversarial boundary forging module typically lie between the normal and abnormal clusters in the projected space.}
    \label{fig:tsne}
\end{figure}

\begin{figure*}[t]
    \centering
    \includegraphics[width=\linewidth]{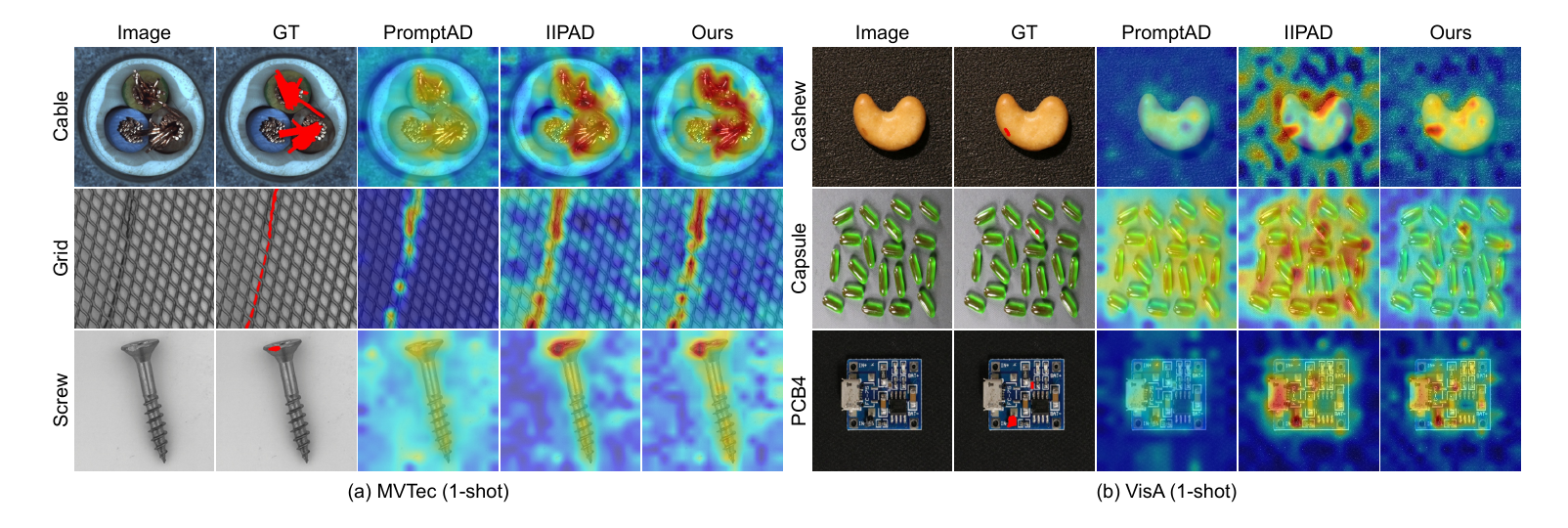}
    \vspace{-15pt}
    \caption{Qualitative anomaly localization results (1-shot). The proposed framework accurately delineates subtle defect boundaries compared to PromptAD and IIPAD.}
    \label{fig:visualization}
\end{figure*}

\paragraph{Qualitative Analysis.}
Figure~\ref{fig:tsne} visualizes the 1-shot feature embeddings to illustrate geometric optimization. Synthetic fence features generated by ABF typically lie between normal and abnormal clusters, aligning with the boundary-regularization objective. Furthermore, Figure~\ref{fig:visualization} shows this optimized boundary yields precise spatial localization, identifying subtle defects often missed by baselines.
\subsection{Ablation Studies}
Ablation studies are conducted under the unified 1-shot setting on MVTec-AD and VisA to evaluate the contributions of the core components.

\begin{table}[htbp]
    \centering
    \small
    \setlength{\tabcolsep}{3pt}
    \caption{Ablation of architectural designs and objective terms under the 1-shot setting. MoE: Universal-Specific Decomposition; Calib.: Query-Aware Semantic Calibration; Deep: Deep Context Refreshing.}
    \label{tab:comprehensive_ablation}
    \begin{tabular}{ccc cc cc cc}
        \toprule
        \multicolumn{3}{c}{DCF Architecture} & \multicolumn{2}{c}{ABF Loss} & \multicolumn{2}{c}{MVTec-AD} & \multicolumn{2}{c}{VisA} \\
        \cmidrule(lr){1-3} \cmidrule(lr){4-5} \cmidrule(lr){6-7} \cmidrule(lr){8-9}
        MoE & Calib. & Deep & $\mathcal{L}_{\text{bal}}$ & $\mathcal{L}_{\text{div}}$ & AUC$_I$ & AUC$_P$ & AUC$_I$ & AUC$_P$ \\
        \midrule
        % DCF Ablation
        $\times$   & $\times$   & $\times$   & \checkmark & \checkmark & 93.4 & 94.3 & 84.9 & 95.2 \\
        \checkmark & $\times$   & $\times$   & \checkmark & \checkmark & 94.2 & 95.1 & 86.1 & 96.2 \\
        \checkmark & \checkmark & $\times$   & \checkmark & \checkmark & 94.5 & 95.7 & 86.7 & 96.8 \\
        \checkmark & \checkmark & \checkmark & \checkmark & \checkmark & \textbf{94.8} & \textbf{96.2} & \textbf{87.3} & \textbf{97.4} \\
        \midrule
        % ABF Ablation
        \checkmark & \checkmark & \checkmark & $\times$   & $\times$   & 92.7 & 95.3 & 84.9 & 96.2 \\
        \checkmark & \checkmark & \checkmark & $\times$   & \checkmark & 92.5 & 94.7 & 84.7 & 96.0 \\
        \checkmark & \checkmark & \checkmark & \checkmark & $\times$   & 94.4 & 95.7 & 87.0 & 96.8 \\
        \bottomrule
    \end{tabular}
\end{table}

\begin{table}[htbp]
    \centering
    \small
    \setlength{\tabcolsep}{5pt}
    \caption{Computational efficiency (1-shot) evaluated on MVTec-AD using a single RTX 4090. All methods utilize a ViT-L/14 backbone and operate under the unified multi-class setting. The training time reports the end-to-end wall-clock time for 20 epochs at $224{\times}224$ resolution, incorporating a 10-step feature-space PGD for the proposed method.}
    \label{tab:efficiency_comparison}
    \begin{tabular}{l c cc cc}
        \toprule
        \multirow{2.5}{*}{Method} & \multirow{2.5}{*}{Total Params} & \multicolumn{2}{c}{VRAM (GB)} & \multicolumn{2}{c}{Time \& Speed} \\
        \cmidrule(lr){3-4} \cmidrule(lr){5-6}
        & & Train & Infer. & Train (Min) & Infer. (FPS) \\
        \midrule
        PromptAD~\cite{li2024promptad} & 428M & \textbf{2.8} & 3.5 & 3.0 & \textbf{12.4} \\
        IIPAD~\cite{lv2025one} & 987M & 24.0 & 10.0 & 3.2 & 8.2 \\
        \midrule
        \textbf{ABounD (Ours)} & 468M & 3.6 & \textbf{2.5} & \textbf{2.5} & 11.7 \\
        \bottomrule
    \end{tabular}
\end{table}

\paragraph{Analysis of Dynamic Concept Fusion.}
Table~\ref{tab:comprehensive_ablation} (top) demonstrates the impact of dynamic concept fusion. Removing it entirely ($\times, \times, \times$) forces reliance on static prompts, yielding the lowest performance. Introducing universal-specific decomposition (MoE) provides a semantic foundation, improving accuracy. Query-aware semantic calibration (Calib.) adapts prompts to individual visual queries, reducing instance-level variations. Deep context refreshing (Deep) maintains these representations across layers, ensuring a stable semantic anchor. This combination highlights the necessity of hierarchical prompt adaptation.
\paragraph{Analysis of Adversarial Boundary Forging.}
Table~\ref{tab:comprehensive_ablation} (bottom) evaluates adversarial boundary forging. Disabling it ($\mathcal{L}_{\text{bal}} \times$, $\mathcal{L}_{\text{div}} \times$) degrades performance, showing semantic anchors alone yield ambiguous boundaries. Applying only dispersion loss ($\mathcal{L}_{\text{div}}$) hurts performance, indicating undirected feature diversity is ineffective. Conversely, using only balance loss ($\mathcal{L}_{\text{bal}}$) locates the geometric margin, improving results. Combining both achieves the highest performance: $\mathcal{L}_{\text{bal}}$ positions the boundary while $\mathcal{L}_{\text{div}}$ regularizes it with distributed adversarial examples.
\paragraph{Justification of Core Mechanisms.}
To ensure improvements stem from architectural design rather than mere capacity or variance increases, we evaluate the mechanisms against standard alternatives. Table~\ref{tab:core_mechanisms} compares dynamic concept fusion against a prompt pool with identical capacity. Scaling parameters via a standard pool yields marginal gains, whereas our structural disentanglement prevents multi-class feature space collapse. Furthermore, directed adversarial search is compared against unstructured noise. Random noise uniformly expands the normal manifold, yielding unstable image-level gains and encroaching upon adjacent classes. In contrast, our boundary optimization locates semantic frontiers, yielding consistent improvements across both metrics.
\begin{table}[htbp]
    \centering
    \small
    \setlength{\tabcolsep}{4pt}
    \caption{Justification of core mechanisms against standard alternatives (1-shot setting). The evaluation confirms the necessity of structural disentanglement over parameter scaling, and directed boundary search over unstructured noise.}
    \label{tab:core_mechanisms}
    \begin{tabular}{l c c c c}
        \toprule
        \multirow{2}{*}{Method Configuration} & \multicolumn{2}{c}{MVTec-AD} & \multicolumn{2}{c}{VisA} \\
        \cmidrule(lr){2-3} \cmidrule(lr){4-5}
        & AUROC$_I$ & PRO & AUROC$_I$ & PRO \\
        \midrule
        \multicolumn{5}{l}{\textit{Ablation on DCF Capacity and Architecture}} \\
        Static Prompt (w/o MoE) & 93.4 & 88.5 & 84.9 & 85.1 \\
        Same-size Prompt Pool & 93.9 & 89.6 & 85.8 & 86.8 \\
        \textbf{Ours (DCF: MoE + Routing)} & \textbf{94.8} & \textbf{91.1} & \textbf{87.3} & \textbf{88.6} \\
        \midrule
        \multicolumn{5}{l}{\textit{Ablation on Feature Perturbation Strategies}} \\
        Base (No Feature Perturbation) & 92.7 & 88.8 & 84.9 & 87.3 \\
        + Gaussian Noise & 93.1 & 90.8 & 85.3 & 88.2 \\
        + Uniform Noise & 93.0 & 90.6 & 85.2 & 88.1 \\
        \textbf{+ Ours (PGD-based ABF)} & \textbf{94.8} & \textbf{91.1} & \textbf{87.3} & \textbf{88.6} \\
        \bottomrule
    \end{tabular}
\end{table}
\paragraph{Efficiency and Scalability.} 
Table~\ref{tab:efficiency_comparison} details computational efficiency. While all methods use a 427M ViT-L backbone, our 41M module results in a total parameter count (468M) lower than IIPAD (987M). This enables rapid convergence, completing 1-shot training in 2.5 minutes using 3.6 GB VRAM, avoiding IIPAD's memory bottleneck (24.0 GB). During deployment, it requires 2.5 GB of memory at 11.7 FPS, demonstrating suitability for resource-constrained environments.

\section{Conclusion}
This work proposes an adversarial boundary-driven framework for multi-class few-shot anomaly detection. By combining dynamic concept fusion for semantic anchoring with adversarial boundary forging for decision margin optimization, the proposed method creates a discriminative feature space. Experiments on seven benchmarks demonstrate that the framework consistently outperforms existing unified few-shot methods while maintaining low computational overhead.
% \section*{Acknowledgements}
% Please insert your acknowledgments here.

\bibliographystyle{splncs04}
\bibliography{main}

\end{document}